\documentclass[conference]{IEEEtran}
\IEEEoverridecommandlockouts
\usepackage{cite}
\usepackage{amsmath,amssymb,amsfonts,upquote,textcomp,multirow}
\usepackage{caption}
\usepackage{subcaption}
\usepackage[ruled,vlined]{algorithm2e}
\usepackage{algorithmic}
\usepackage{graphicx}
\usepackage{textcomp}
\usepackage{xcolor}
\usepackage{booktabs}

\usepackage{tikz}
\usetikzlibrary{fit,calc}

\colorlet{pink}{red!40}
\colorlet{blue}{cyan!60}

\def\BibTeX{{\rm B\kern-.05em{\sc i\kern-.025em b}\kern-.08em
    T\kern-.1667em\lower.7ex\hbox{E}\kern-.125emX}}

\usepackage{scalerel}
\usepackage{tikz}
\usetikzlibrary{svg.path}

\definecolor{orcidlogocol}{HTML}{A6CE39}
\tikzset{
  orcidlogo/.pic={
    \fill[orcidlogocol] svg{M256,128c0,70.7-57.3,128-128,128C57.3,256,0,198.7,0,128C0,57.3,57.3,0,128,0C198.7,0,256,57.3,256,128z};
    \fill[white] svg{M86.3,186.2H70.9V79.1h15.4v48.4V186.2z}
                 svg{M108.9,79.1h41.6c39.6,0,57,28.3,57,53.6c0,27.5-21.5,53.6-56.8,53.6h-41.8V79.1z M124.3,172.4h24.5c34.9,0,42.9-26.5,42.9-39.7c0-21.5-13.7-39.7-43.7-39.7h-23.7V172.4z}
                 svg{M88.7,56.8c0,5.5-4.5,10.1-10.1,10.1c-5.6,0-10.1-4.6-10.1-10.1c0-5.6,4.5-10.1,10.1-10.1C84.2,46.7,88.7,51.3,88.7,56.8z};
  }
}

\newcommand\orcidicon[1]{\href{https://orcid.org/#1}{\mbox{\scalerel*{
\begin{tikzpicture}[yscale=-1,transform shape]
\pic{orcidlogo};
\end{tikzpicture}
}{|}}}}

\PassOptionsToPackage{hyphens}{url}
\usepackage{hyperref}

\begin{document}

\title{TinyReptile: TinyML with Federated Meta-Learning}


\author{\IEEEauthorblockN{1\textsuperscript{st} Haoyu Ren\textsuperscript{\orcidicon{0000-0002-0241-6507}} \,}
\IEEEauthorblockA{\textit{Siemens AG} \\
\textit{Technical University of Munich}\\
Munich, Germany \\
haoyu.ren@siemens.com}
\and
\IEEEauthorblockN{2\textsuperscript{nd} Darko Anicic\textsuperscript{\orcidicon{0000-0002-0583-4376}}}
\IEEEauthorblockA{\textit{Siemens AG} \\
Munich, Germany \\
darko.anicic@siemens.com}
\and
\IEEEauthorblockN{3\textsuperscript{rd} Thomas A. Runkler\textsuperscript{\orcidicon{0000-0002-5465-198X}}}
\IEEEauthorblockA{\textit{Siemens AG} \\
\textit{Technical University of Munich}\\
Munich, Germany \\
thomas.runkler@siemens.com}
}

\maketitle

\begin{abstract}
Tiny machine learning (TinyML) is a rapidly growing field aiming to democratize machine learning (ML) for resource-constrained microcontrollers (MCUs). Given the pervasiveness of these tiny devices, it is inherent to ask whether TinyML applications can benefit from aggregating their knowledge. Federated learning (FL) enables decentralized agents to jointly learn a global model without sharing sensitive local data. However, a common global model may not work for all devices due to the complexity of the actual deployment environment and the heterogeneity of the data available on each device. In addition, the deployment of TinyML hardware has significant computational and communication constraints, which traditional ML fails to address. Considering these challenges, we propose TinyReptile, a simple but efficient algorithm inspired by meta-learning and online learning, to collaboratively learn a solid initialization for a neural network (NN) across tiny devices that can be quickly adapted to a new device with respect to its data. We demonstrate TinyReptile on Raspberry Pi 4 and Cortex\mbox-M4 MCU with only 256-KB RAM. The evaluations on various TinyML use cases confirm a resource reduction and training time saving by at least two factors compared with baseline algorithms with comparable performance.
\end{abstract}



\section{Introduction}

In the past decade, the evolvement of machine learning (ML) applications has been driven by the advent of big data and increased processing capability. As a result, large-scale artificial intelligence (AI) models have been developed for improved performance, requiring enormous computational resources and massive power consumption. For example, Microsoft and NVIDIA have introduced an ML model named “Megatron-Turing NLG”~\cite{Smith2022} with 530 billion neurons that uses
approximately 5 GWh of electricity to train. The community is increasingly aware that deploying an advanced ML model is costly and unsustainable. 

Given these concerns, Tiny machine learning (TinyML) has risen to popularity at the intersection of ML and embedded systems, shifting data processing from data centers to ever-smaller Internet of Things (IoT) units. These low-cost devices typically run on batteries and are designed to operate for extended periods using limited resources, e.g., 256-KB RAM and 64-MHz CPUs. TinyML enables edge AI directly on IoT devices in near real-time as close to where the action occurs. By avoiding data transmission to the cloud, TinyML brings advantages in terms of data privacy, latency, and energy efficiency. It is estimated that more than 250 billion embedded devices are in use today, with a steadily growing demand, especially in industry~\footnote{\url{https://venturebeat.com/ai/why-tinyml-is-a-giant-opportunity/}}. Given the vast number of IoT devices deployed in production environments, an important research direction is to share and integrate insights learned across tiny devices to improve the performance and robustness of TinyML applications.


Federated learning (FL) offers a method for learning a global model on distributed agents/devices. FL can protect local data privacy from design because only model updates are merged into the global model, and the raw data are never copied to the cloud. However, IoT devices are distributed over the field, and the real-world environment is complex and constantly changing, making local data generally not uniform across devices. The environments of two even neighboring devices can differ significantly. Besides, each device may have its requirements for the ML task, e.g., different output classes of interest. Therefore, a common shared model trained by FL may perform arbitrarily poorly when applied to a local device. Moreover, most FL algorithms are designed to run on powerful machines without considering computational resource constraints. Their communication schemas transmit weight updates to the central server on a carefully scheduled and consistent basis. All these are not feasible for many TinyML applications because IoT devices are highly resource-constrained in terms of communication, memory, and computation. An unstable network connection may prevent communication for long periods, and insufficient on-device resources may hinder the operations of these algorithms. 

To tackle these challenges, we present TinyReptile, a modified formulation of the well-known Reptile algorithm~\cite{Nichol2018}, to perform model-agnostic meta-learning in a federated setting across tiny devices by leveraging online learning. Meta-learning intends to find an initial shared model that current or new agents can quickly adapt to their environments by performing only a few steps of gradient descent with respect to a small amount of local data. In other words, meta-learning is about learning to learn. Our TinyReptile algorithm is based on the concept of online learning, where we design the algorithm to enable incremental on-device learning in a distributed manner while considering resource constraints. With online learning, we run the model on the data as they arrive in a streaming fashion without storing historical data, which is close to reality in many industrial applications. TinyReptile also retains the benefits of a serial schema in that devices can communicate with the central server one after the other without relying on a consistent and concurrent connection. Furthermore, it contributes to a more customized model for each device. 

By comparing TinyReptile with FedAVG and FedSGD~\cite{McMahan2016}, we show that traditional FL architectures can fail when devices exhibit heterogeneity in local data. We investigate TinyReptile on two established and one constructed meta-learning datasets, namely, “Sine-wave example” for a regression problem, “Omniglot”~\cite{Lake2019} for an image classification task, and our “Keywords spotting” for an audio classification task. Because most meta-learning benchmark datasets are designed for image classification purposes, we contribute the “Keywords spotting” dataset to evaluate meta-learning on an audio classification problem, which is created from the popular TinyML dataset “Speech commands”~\cite{Warden2018}. We compare TinyReptile with the baseline algorithm, namely, Reptile, on Raspberry Pi and Arduino microcontrollers (MCUs) across these datasets. Evaluation results show that TinyReptile can reduce computational resources and the training time by a factor of at least two compared with the baseline Reptile algorithm while achieving comparable performance. To guide the deployment of TinyReptile on tiny hardware, we characterize the hyperparameters in TinyReptile under various settings and present the results in Appendix~\ref{appendix:A}.

The remainder of this paper discusses related work on TinyML, online learning, FL, and meta-learning in Section~\ref{section:related_work}. Section~\ref{section:approach} demonstrates TinyReptile and its methodology. Section~\ref{section:experiments} describes the benchmarking datasets and the experimental settings and analyzes the results. Finally, section~\ref{section:conclusion} concludes the paper and discusses future work.

\section{Related Work}
\label{section:related_work}

\subsubsection*{TinyML}
Recently, we have witnessed significant progress in TinyML across research and industry. Reviews~\cite{Dutta2021}~\cite{Ray2022} provide an intuitive overview of the different directions of TinyML, along with challenges and opportunities. In particular, the related research can be described in three main aspects: hardware, software, and applications. For hardware, novel technologies such as analog in-memory computing~\cite{Datta2022} have been proposed to optimize processing capabilities in resource-frugal devices. Deep-learning accelerators~\cite{Scherer2022} and intelligent sensors~\cite{Lu2022} are specialized to execute AI workloads efficiently. Software advances are led by the development of compressing methods~\cite{Zhuo2022}, management systems~\cite{Ren2022}, and efficient algorithms~\cite{Szydlo2022}. However, most available solutions, such as TensorFlow Lite for MCUs~\cite{David2020}, only support the deployment of static ML models for inference, which prevents devices from learning the newest knowledge from the field. Only a few studies~\cite{Ravaglia2021}~\cite{Lin2020} have discussed the possibilities of model training on tiny devices. Nevertheless, the advancement of TinyML has started to benefit society, such as in predictive maintenance~\cite{Njor2022} and healthcare~\cite{Tsoukas2021}.

\subsubsection*{Federated Learning}
FL focuses on joint training of a global model under the orchestration of a central server while keeping training data decentralized. Research on FL first emerged in 2016~\cite{McMahan2016}, when the two gold standard baseline algorithms, “FedAVG” and “FedSGD,” were proposed. Since then, continued works have been proposed in various areas: optimization algorithms~\cite{Li2020}, model update compression~\cite{Sattler2020}, differential privacy~\cite{Jiang2021}, and robustness~\cite{Aramoon2021}. Some FL studies have been conducted on edge devices with lower capacities, such as Raspberry Pi~\cite{Gao2020}. However, little effort has been devoted to applying FL on highly constrained devices in the context of TinyML. The current state of the art is restricted to two related works~\cite{Kopparapu2021}~\cite{Grau2021}. Their experiments are implemented either in simulation or in a controlled environment on a small scale, which cannot reflect situations in the real world. For example, FL suffers from client heterogeneity, which can lead to slow and unsatisfactory learning progress.

\subsubsection*{Meta-learning} 
Meta-learning is proposed for fast local adaptation with small datasets, making it an ideal candidate for TinyML scenarios. Meta-learning is similar to transfer learning. In transfer learning, a global model can be pretrained and later fine-tuned on a specific small dataset, but with no guarantee of learning a good initialization for generalization~\cite{Hospedales2021}. Conversely, meta-learning trains a common model with the explicit objective of being easily fine-tuned. The very first advance in meta-learning, called “model-agnostic meta-learning” (MAML)~\cite{Finn2017}, brings the concept to light through gradient-based optimization. This is followed by several papers studying its convergence behavior and empirical properties~\cite{Raghu2019}~\cite{Jiang2019}. However, MAML requires the computation of higher-order derivatives, making it cumbersome and computationally intensive. A few lightweight variants with simpler operations are proposed to alleviate this problem, including First Order MAML and Reptile~\cite{Nichol2018}. Besides, meta-learning has been investigated in federated~\cite{Fallah2020}~\cite{Lin2020a} and online learning settings~\cite{Finn2019}~\cite{Acar2021}. However, to the best of our knowledge, no relevant work has focused on applying meta-learning to constrained devices. 

\subsubsection*{Online learning} 
An ML system has two tasks: inference and training. Online ML involves performing these tasks online, i.e., processing data sequentially without revisiting previous samples, saving memory resources, and ideally not sacrificing model performance. However, less attention is paid to online learning~\cite{Gomes2019}~\cite{Gepperth2016} compared with batch/offline learning because most ML engineers tend to assume that devices and their data are always available as a batch. River is a popular Python library for online ML~\cite{Montiel2021}. We believe that online learning is a perfect fit for TinyML applications. The concept helps us develop algorithms with lower memory footprints and keep models up-to-date in changing environments. Interesting works have been devoted to this area~\cite{Ravaglia2021a}~\cite{Ren2021}. In this study, we integrate online learning into meta-learning to enable meta-learning on constrained devices in a federated setting.

\section{Approach}
\label{section:approach}

The objective of meta-learning is to teach a neural network (NN) how to adapt or extrapolate quickly to new tasks (environments) with only a few training examples. We want to achieve this objective by leveraging FL across heterogeneous and constrained IoT devices, each of which has limited exposure to labeled samples during training. In this section, we define the problem setup and present the formulation of TinyReptile.

\subsection{Meta-learning}
First, we describe the optimization problem of meta-learning. Here we consider a distribution of independent tasks $T$ that share a similarity, e.g., all tasks use an NN of the same structure for image classification. However, every task has a different output of interest, e.g., one is to classify “dog vs. cat,” and another is to classify “apple vs. pear.” We aim to find good initial weights $\phi$ for the NN that can quickly learn when given a new task $t$ sampled from the distribution $T$ that has never been encountered during training, e.g., the classification of “bicycle vs. motorcycle.” 

We divide the tasks into training tasks $T_{training}$ and testing tasks $T_{testing}$. The meta-learning algorithm $F$ trains the NN on $T_{training}$ and evaluates the quality of the trained model using $T_{testing}$. Each task $t$ has a dataset $D$ split into two parts: a support set $S$ for training within the task, and a query set $Q$ for testing within the task. $D = \langle S, Q \rangle$ typically has a limited number of samples. To evaluate a trained initialization, we first need to fine-tune the model weights $\phi$ on the support set $S$ across all $T_{testing}$, each for $k$ steps, respectively. Then, the fine-tuned new weights $\hat{\phi}$ are tested on the corresponding query set $Q$ of each testing task, and finally, the results for all $T_{testing}$ are averaged. Because meta-learning aims to find a fast learner, the training step $k$ is typically defined as a small number. The goal of the meta-learning algorithm $F$ is to minimize the loss function across the tasks:
\begin{equation}
    L(\phi) = \sum\limits_{n=1}^N l_n(\hat{\phi}^k_n),
    \label{e1}
\end{equation}
where $\hat{\phi}^k_n$ denotes the model weights fine-tuned for $k$ steps from the initial weights $\phi$ on the support set of the task $n$, $l_n(\hat{\phi}^k_n)$ denotes the loss on the query set of the task $n$, and $N$ represents the number of tasks.

One can distinguish the goal of meta-learning from transfer learning, where meta-learning finds $\phi$ that can perform well after a quick fine-tuning on each new task. In other words, meta-learning seeks the potential performance of a model. On the other hand, the loss function of transfer learning is defined as follows: 
\begin{equation}
    L(\phi) = \sum\limits_{n=1}^N l_n(\phi_n),
\end{equation}
which indicates that transfer learning aims to find a good initialization that works on all tasks without fine-tuning. Namely, transfer learning looks for the current performance of a model.

\subsection{TinyReptile}

We present our federated meta-learning algorithm for TinyML called TinyReptile. TinyReptile learns an initialization for an NN by leveraging FL across heterogeneous IoT devices. The local fine-tuning should be fast when we optimize the initialization for a new ML problem based on only a few labeled local examples. TinyReptile is an online learning version of Reptile tailored for TinyML. By leveraging the characteristics of online learning, TinyReptile processes data sequentially as they arrive, saving computing resources and enabling inference and training directly on constrained IoT devices. 

Notably, in meta-learning, there are tasks with their datasets, whereas, in TinyReptile, we have devices that possess their datasets or sensors to collect data. Building on the concept of FL, TinyReptile minimizes a version of the loss function~\ref{e1} by representing the ML problem of each device as a training task in meta-learning. TinyReptile improves the adaptation ability of the initial model across all tasks in the distributed devices $T$:
\begin{equation}
    \mathop{minimize}_{\phi} \mathbb{E}_t [ \frac{1}{2} L(\hat{\phi}^k_t, \phi^*_t) ]^2,
\end{equation}
where $\phi$ denotes the initialization, and $L$ represents the distance between the fine-tuned weights $\hat{\phi}^k_t$ and the optimal weights $\phi^*_t$ for device $t$. The TinyReptile algorithm is defined as follows (we highlight the major differences between TinyReptile and Reptile in orange):

\begin{algorithm}[htbp]
  \caption{TinyReptile}
  \begin{algorithmic}[1]
    \STATE Central model weights $\phi$;
    \STATE Central server learning rate $\alpha$;
    \STATE Clients $T = \langle T_{training}, T_{testing} \rangle$, each with local \textcolor{orange}{streaming} data $D$. For evaluation, $D$ on $T_{testing}$ is divided into support set (S) and query set (Q), $D = \langle S, Q \rangle$;
    \STATE Client learning rate $\beta$;
    \FOR{round = 1, 2, ...} 
        \STATE Sample \textcolor{orange}{one} client $t$ from $T_{training}$ with respect to \newline the loss $L_t$;
        \STATE Send $\phi$ to the client $t$;
        
        \FOR{\textcolor{orange}{each $x$} in the local \textcolor{orange}{streaming} support set $S_t$} 
            \STATE Compute $\hat{\phi}^{k+1}_t = SGD(\hat{\phi}^k_t, x, \beta, L_t)$, denoting \newline \textcolor{orange}{one} step of SGD on the client;
        \ENDFOR

        \STATE Send $\hat{\phi}_t$ back to the server;
        \STATE Update the central model weights: \hspace*{5mm}
         $\phi \leftarrow  \phi + \alpha (\hat{\phi}_t - \phi)$
         \STATE (optional) Evaluate $\phi$ on $T_{testing}$;
    \ENDFOR
  \end{algorithmic}
  \label{a1}
\end{algorithm}

\begin{figure*}[tbp]
     \centering
     \begin{subfigure}[b]{0.325\textwidth}
         \centering
         \includegraphics[width=\textwidth]{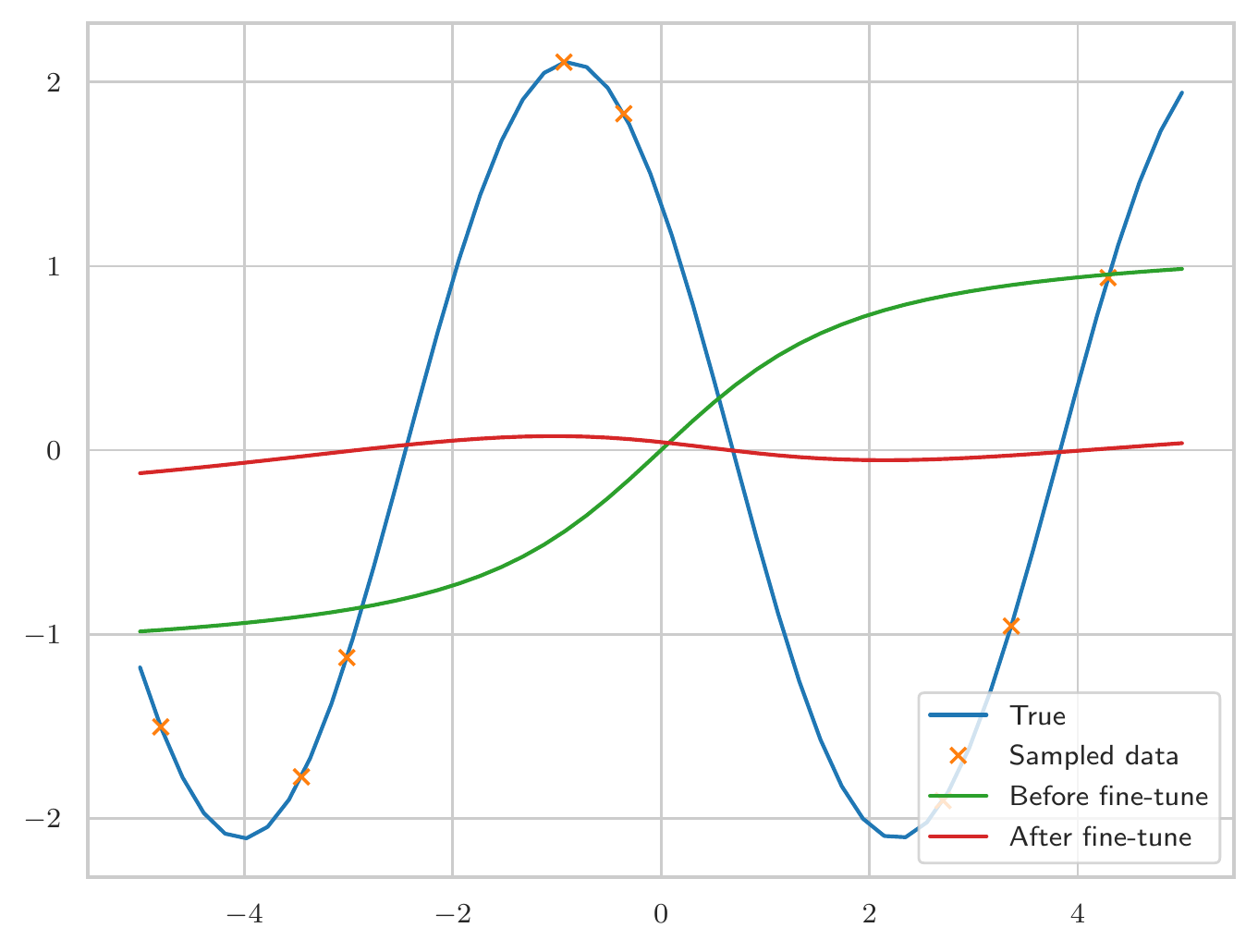}
         \caption{Transfer learning}
         \label{fig:sine_compare_fedSGD}
     \end{subfigure}
     \hfill
     \begin{subfigure}[b]{0.325\textwidth}
         \centering
         \includegraphics[width=\textwidth]{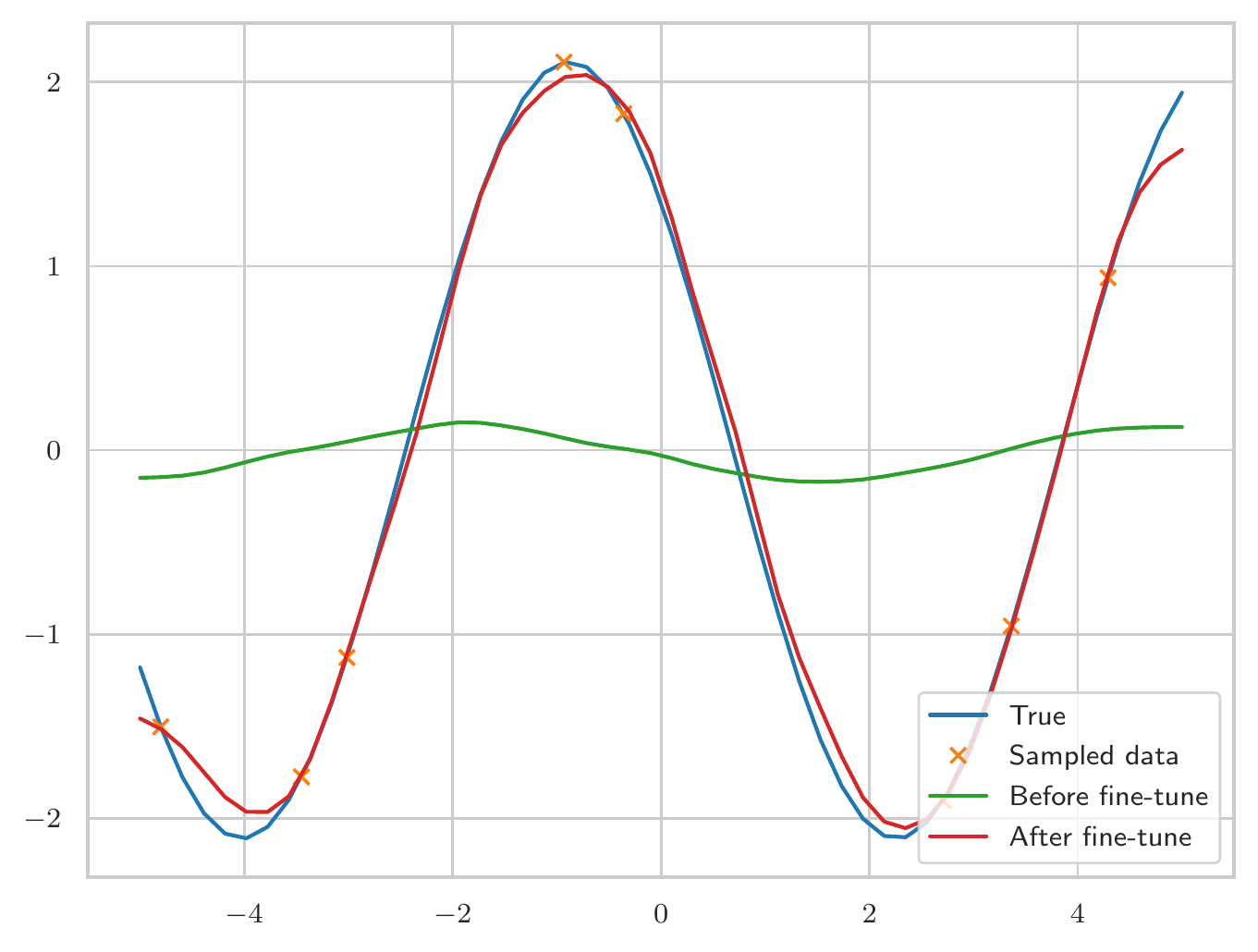}
         \caption{Reptile.}
         \label{fig:sine_compare_reptile}
     \end{subfigure}
     \hfill
     \begin{subfigure}[b]{0.325 \textwidth}
         \centering
         \includegraphics[width=\textwidth]{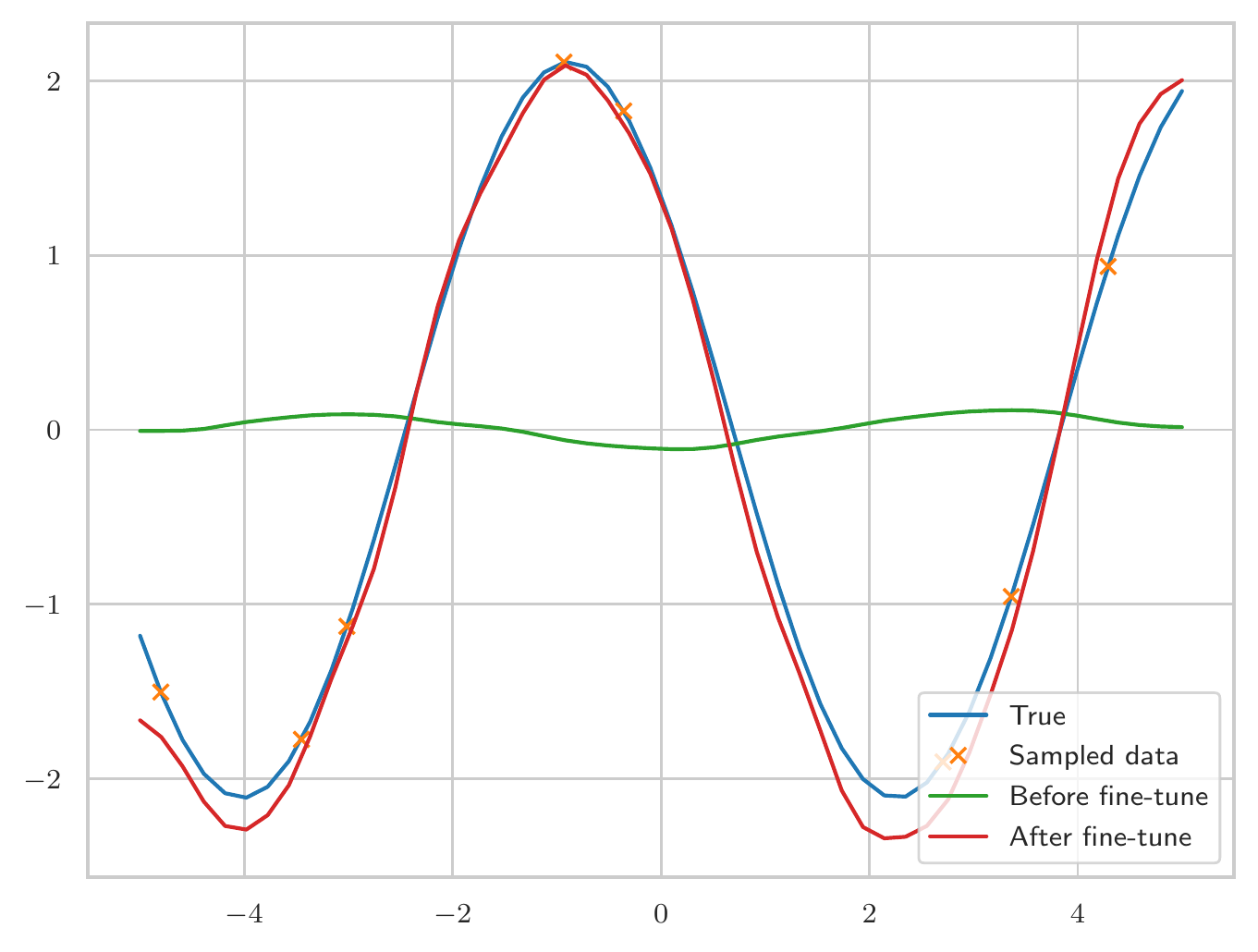}
         \caption{TinyReptile (ours).}
         \label{fig:sine_compare_tinyReptile}
     \end{subfigure}
        \caption{Demonstration of transfer learning, Reptile, and TinyReptile on the Sine-wave regression example, where we fine-tune the trained NNs on the support set (eight sampled points) of a testing client for eight local SGD updates. The model consists of four fully connected layers: 1 → 32 → 32 → 1. This shows that the NNs trained with Reptile and TinyReptile can quickly converge to the sampled sine wave and derive values away from the sampled points, which is difficult with transfer learning.}
        \label{fig:1}
\end{figure*}

With online learning, TinyReptile learns incrementally. As mentioned above, our algorithm is a modified formulation of the Reptile algorithm [5]. The critical difference between our method and Reptile is that Reptile performs batch training on each client, where the NN is trained on the entire support set in a batched manner, and the training data are stacked and reused, which is resource-demanding. Our method processes one data point at a time in a streaming process, and that data can be discarded after each update, which is highly resource-efficient. In each round, TinyReptile performs stochastic gradient descent (SGD) sequentially against the streaming data from the support set $S$ of a sampled training client, e.g., real-time sensor data of an IoT device. Then, the initial central weights are updated toward the weights obtained after the SGD update. Next, the trained initialization can be assessed by the testing clients if evaluation is required. In this manner, TinyReptile attempts to move the initial weights closer to an optimal point nearest to all tasks/devices. Thus, TinyReptile learns an NN initialization that can be quickly fine-tuned on a new task/device. In other words, TinyReptile optimizes for generalization. Interested readers can find more theoretical analysis in the Reptile publication~\cite{Nichol2018}. The key advantages of TinyReptile are related to the following:


\begin{itemize}
\item Robust: TinyReptile has a serial communication schema. Many FL algorithms apply parallel communication schemas, requiring all clients to be stably connected to a central server during training, and the processing capabilities of all devices must be similar, which is unrealistic in many real-world production environments. A disturbance in the network or some “stragglers” among the clients may affect or even interrupt the operation. In contrast, TinyReptile evaluates one device in each round, facilitating the stability of the learning process, especially in the TinyML environment, where many IoT devices are deployed in remote areas with limited network coverage. 

\item Resource-efficient: on-device learning is difficult on constrained devices because they typically do not have sufficient resources for saving and processing large amounts of data. With online learning, data arrive in a stream and are processed one after another in TinyReptile. Data can be effectively discarded after each learning iteration. At any time, only one data sample lives in memory, and we do not need to store historical data. Compared with other offline/batch learning algorithms, TinyReptile requires significantly fewer resources.

\item Scalable: The resource efficiency of TinyReptile makes its core compatible with various devices, from powerful to tiny ones. Owing to its serial communication schema, any device running TinyReptile can join and exit the learning process anytime, making the algorithm scalable and reliable in production. 
\end{itemize}

\begin{table}[tbp]
\centering
\caption{Overview of the models used in the experiments.}
\resizebox{\columnwidth}{!}{%
\begin{tabular}{@{}llll@{}}
\toprule
                           & Model Type & Size & Parameters \\ \midrule
Sine-wave example          & Fully connected         & 19.4 KB            & 1153       \\
Keywords spotting (4 Classes) & Convolutional        & 95.7 KB            & 19812   \\
Omniglot (5 Classes)      & Convolutional        & 485.1 KB           & 113733        \\ \bottomrule
\end{tabular}%
}
\label{tab:1}
\end{table}

\section{Experiments and Evaluation}
\label{section:experiments}

This section first briefly describes two popular meta-learning datasets: the Sine-wave regression problem and the Omniglot image classification problem. Then, we introduce our audio classification dataset “Keywords spotting” for meta-learning and explain the motivation for using the dataset. Next, we use the Sine-wave example as a case study to provide an intuitive overview of meta-learning and demonstrate that traditional FL algorithms fail in a meta-learning setting. Afterward, we compare TinyReptile with Reptile by analyzing their real-world performance, in terms of convergence speed, memory requirement, and time consumption, on Raspberry Pi 4 Model 4~\footnote{\url{https://www.raspberrypi.com/products/raspberry-pi-4-model-b/}} and Arduino Nano BLE 33 MCU~\footnote{\url{https://docs.arduino.cc/hardware/nano-33-ble-sense}}. For the experiments, we use the NNs introduced in the MLPerf Tiny benchmark~\cite{Banbury2021} to ensure consistent and comparable results. An overview of these NNs is illustrated in Table.~\ref{tab:1}. Finally, for easy deployment on TinyML devices, the impact of several hyperparameters in TinyReptile is investigated in Appendix~\ref{appendix:A}. We conduct each experiment three times and plot the results as mean ± standard deviation of the measurements.

\subsection{Meta-learning Datasets}

\subsubsection*{Sine-wave}
The Sine-wave regression problem comes from~\cite{Finn2017}, defined as follows: each client/device has an underlying task of fitting a sine function ${f(x) = a \sin(b\, x + c)}$, where the parameters $(a, b, c)$ are chosen randomly. The goal of clients is to collaboratively learn a good NN initialization that can be quickly generalized to a new sine function ${f_t(x) = a_t \sin(b_t\, x + c_t)}$ given only a few sampled $(x_t, y_t)$ pairs.

\subsubsection*{Omniglot}
The Omniglot image dataset contains $C=1623$ characters from 50 different alphabets, with 20 samples for each character. The meta-learning problem is defined as follows: each client $t$ has a classification task of $M$ characters (classes) sampled from $C$. Although all clients have the same number of classes, the characters of each client for the classification task are randomly selected. For example, if $M=5$, all clients train for five classes with labels $0-4$, but each client has different classification characters. The training goal is similar to the Sine-wave problem above: learning good initial weights that can be quickly adapted to a new classification problem with only a few data examples.

\begin{figure}[tbp]
\centerline{\includegraphics[width=1\columnwidth]{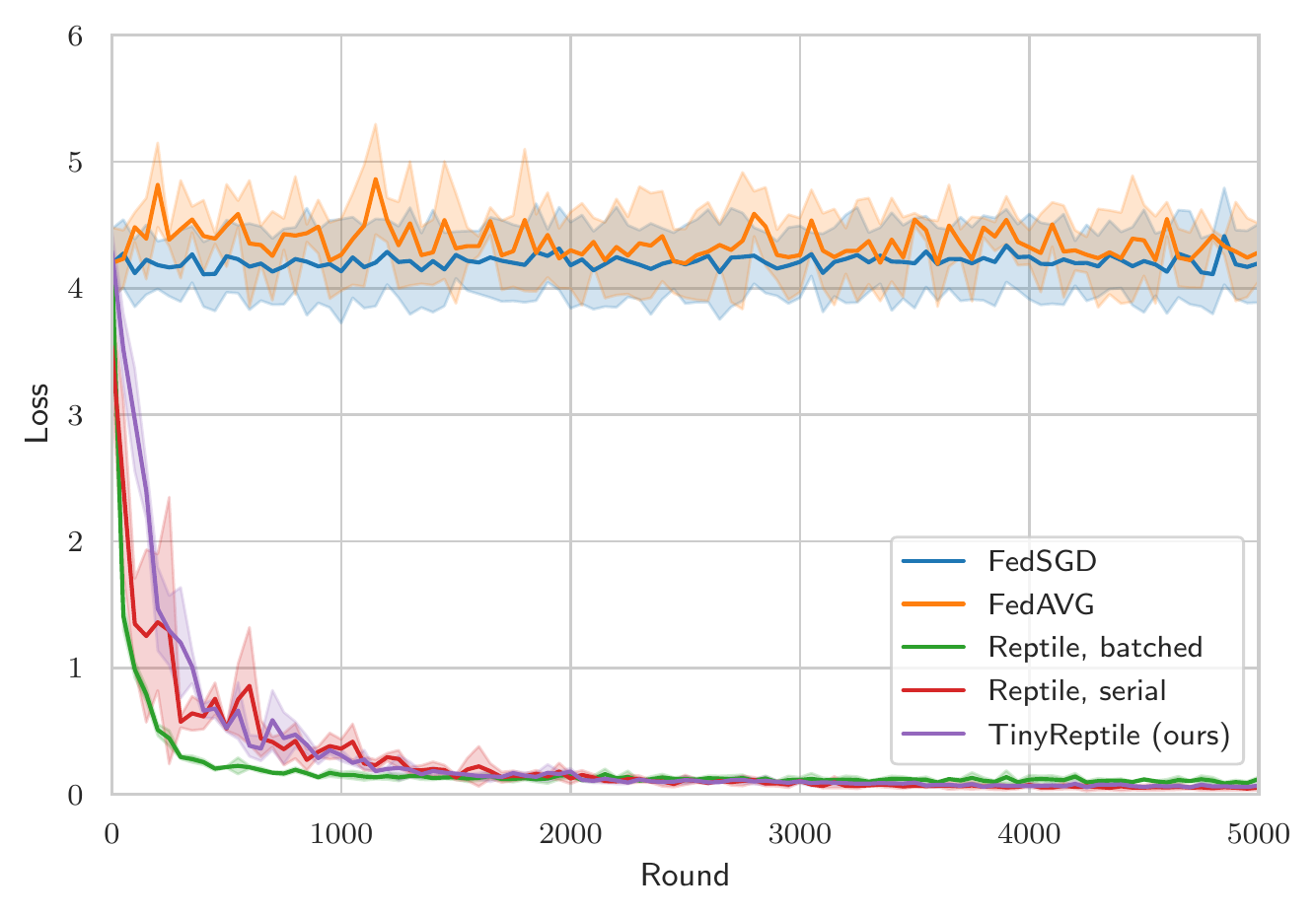}}
\caption{Training convergence of FedSGD, FedAVG, Reptile (batched \& serial), and TinyReptile on the Sine-wave regression example. This shows that our TinyReptile can achieve comparable performance to Reptile. However, it is difficult for traditional FL algorithms, such as FedSGD and FedAVG, to learn meaningful knowledge in a meta-learning setting.}
\label{fig:2}
\end{figure}

\subsubsection*{Keywords spotting}
The “Keywords spotting” audio dataset is a modification of the “Speech command” dataset~\cite{Warden2018}, which contains 35 individual words with more than 1,000 samples for each word. We will not describe the meta-learning setting for this dataset as it is similar to the “Omniglot” classification problem described above but with different classification objectives. Instead, we elaborate on the rationale for proposing this dataset. TinyML is an innovative field that brings ML power to embedded systems, and its potential applications are prolific, ranging from gesture detection to voice recognition. However, we have observed that most established meta-learning datasets are categorized as image classification, which does not fully cover the interests of TinyML. Therefore, we want to contribute a new dataset for evaluating meta-learning on other TinyML use cases. Although the “Speech command” dataset was initially constructed for a different ML problem with limited classes to draw from, we believe that the community can still benefit from the retrofitted “Keywords spotting” dataset.

\subsection{Sine-wave example -- a case study}

We first demonstrate meta-learning on the Sine-wave regression problem by comparing the behavior of transfer learning, Reptile, and our TinyReptile, as illustrated in Fig.~\ref{fig:1}. The results show that transfer learning cannot learn a meaningful initialization in a meta-learning setting. This is because transfer learning aims to train an NN that can handle all sine functions across all clients at once, meaning that a trained model will ideally return the average value of $f(x)$ of all tasks. In the Sine-wave example, the average values of the sine functions $\mathbb{E}_t [f_t (x)]$ are approximated to zero for all $x$ values due to the random parameters $(a, b, c)$ in each task $f(x) = a \sin(b \, x + c)$.

Next, we compare the training progress of FedSGD, \mbox{FedAVG,} Reptile (serial \& batched version), and TinyReptile. The difference between the batched and serial versions of Reptile is that the batched version requires the server to establish connections to multiple clients simultaneously to train the model on them in parallel. In contrast, in the serial version, only one client is connected to the server at any time. As depicted in Fig.~\ref{fig:2}, the results are consistent with our expectation that FedSGD and FedAVG cannot learn a proper initialization when local data exhibit heterogeneity because their learning goals are similar to that of transfer learning. However, our TinyReptile can achieve results comparable to the Reptile algorithm. Although the batched version of Reptile converges faster in the early training phase, it requires more computational resources and concurrent network connections to many clients, which is unsuitable for most TinyML scenarios. 

\subsection{Evaluation Results}

\begin{figure}[tbp]
\centering
\centerline{\includegraphics[width=1\columnwidth]{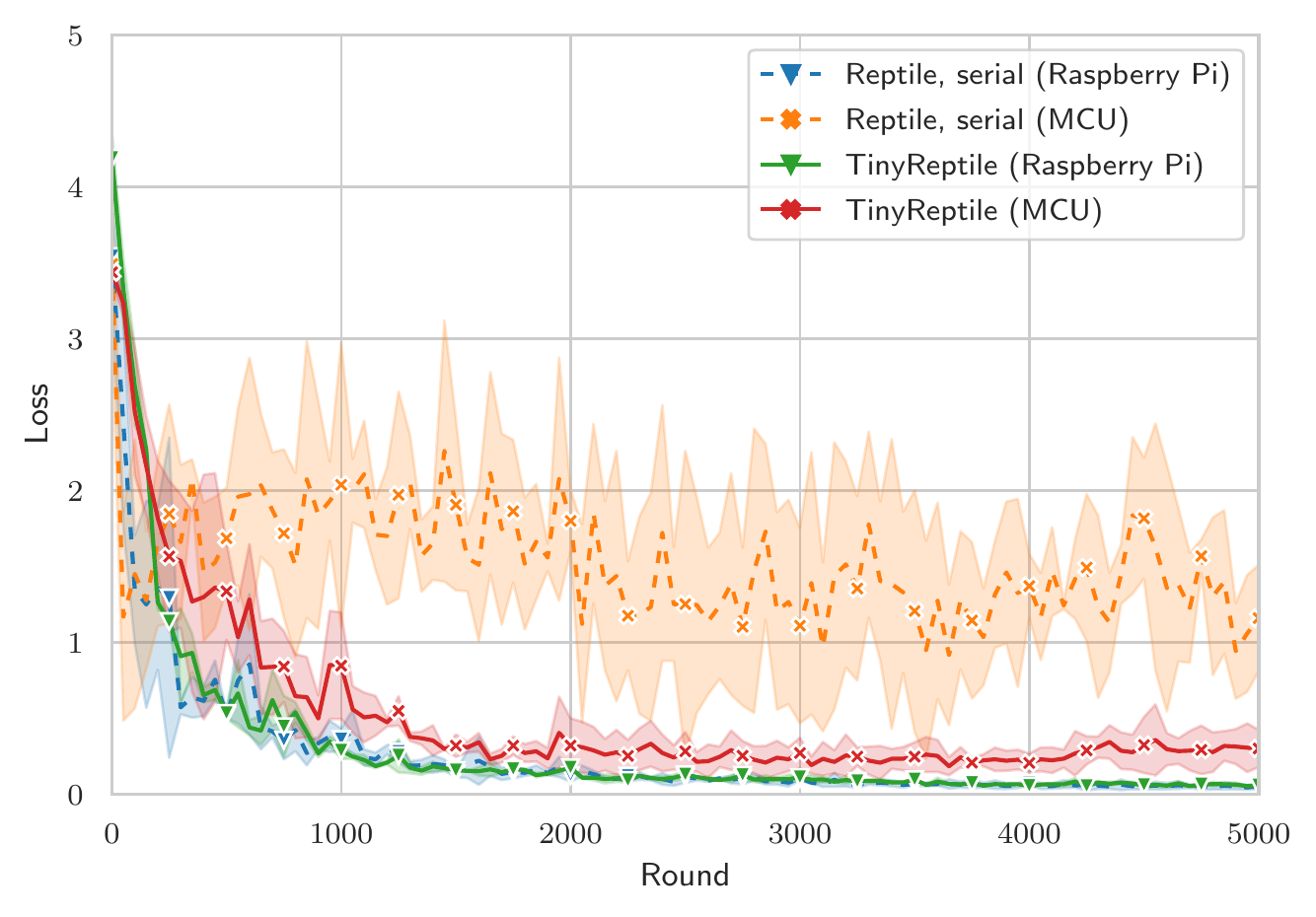}}
\caption{Training convergence of Reptile (serial) and TinyReptile on the Sine-wave regression example on Raspberry Pi 4 and Arduino Nano 33 BLE Sense MCU. This demonstrates that TinyReptile can achieve comparable results to Reptile. The overall performance on the MCU is slightly worse, especially with Reptile, which may be due to the limited numerical precision of the MCU.}
\label{fig:3}
\end{figure}

We evaluate the performance of Reptile and TinyReptile on Raspberry Pi 4 and Arduino Nano BLE 33 MCU. We do not consider other FL algorithms because they are ineffective for meta-learning problems. As mentioned, the NNs we considered in the experiments are described in Table~\ref{tab:1}. We attempted the following ranges of different hyperparameters and found possible combinations that work well for the problems: the client learning rate $\beta$ (0.001--0.02), training steps $k$ (1--32), and support set size $S$ (1--32). However, we did not fine-tune the hyperparameters to optimize the final results. 

\subsubsection*{Memory requirement}

\begin{table}[tbp]
\centering
\caption{Comparison of memory requirement (the results are measured under the support set size $S=32$).}
\resizebox{\columnwidth}{!}{%
\begin{tabular}{@{}lll@{}}
\toprule
                           & \begin{tabular}[c]{@{}l@{}}Reptile\\ (batched \& serial)\end{tabular} & \begin{tabular}[c]{@{}l@{}}TinyReptile\\ (ours)\end{tabular} \\ \midrule
Sine-wave example          & 10.7 KB                     & 4.8 KB               \\
Keywords spotting (4 Classes) & 5816.4 KB                 & 437.2 KB             \\
Omniglot (5 Classes)       & 3778.1 KB                 & 667.2 KB             \\ \bottomrule
\end{tabular}%
}
\label{tab:2}
\end{table}

To run the algorithms on hardware, we need to measure their memory usage for the considered NNs. Notable outcomes are presented in Table.~\ref{tab:2}. Owing to online learning, TinyReptile provides a significant reduction in memory consumption by a factor of at least two across all tasks compared with Reptile. The results also show that only the Sine-wave example can be used for experiments on the Arduino Nano 33 BLE Sense board because this MCU is equipped with only 256 KB of memory.
 
\subsubsection*{Training convergence}

\begin{figure}[tbp]
    \centering
    \subfloat[Omniglot with five output classes.]{
    \includegraphics[width=1\columnwidth]{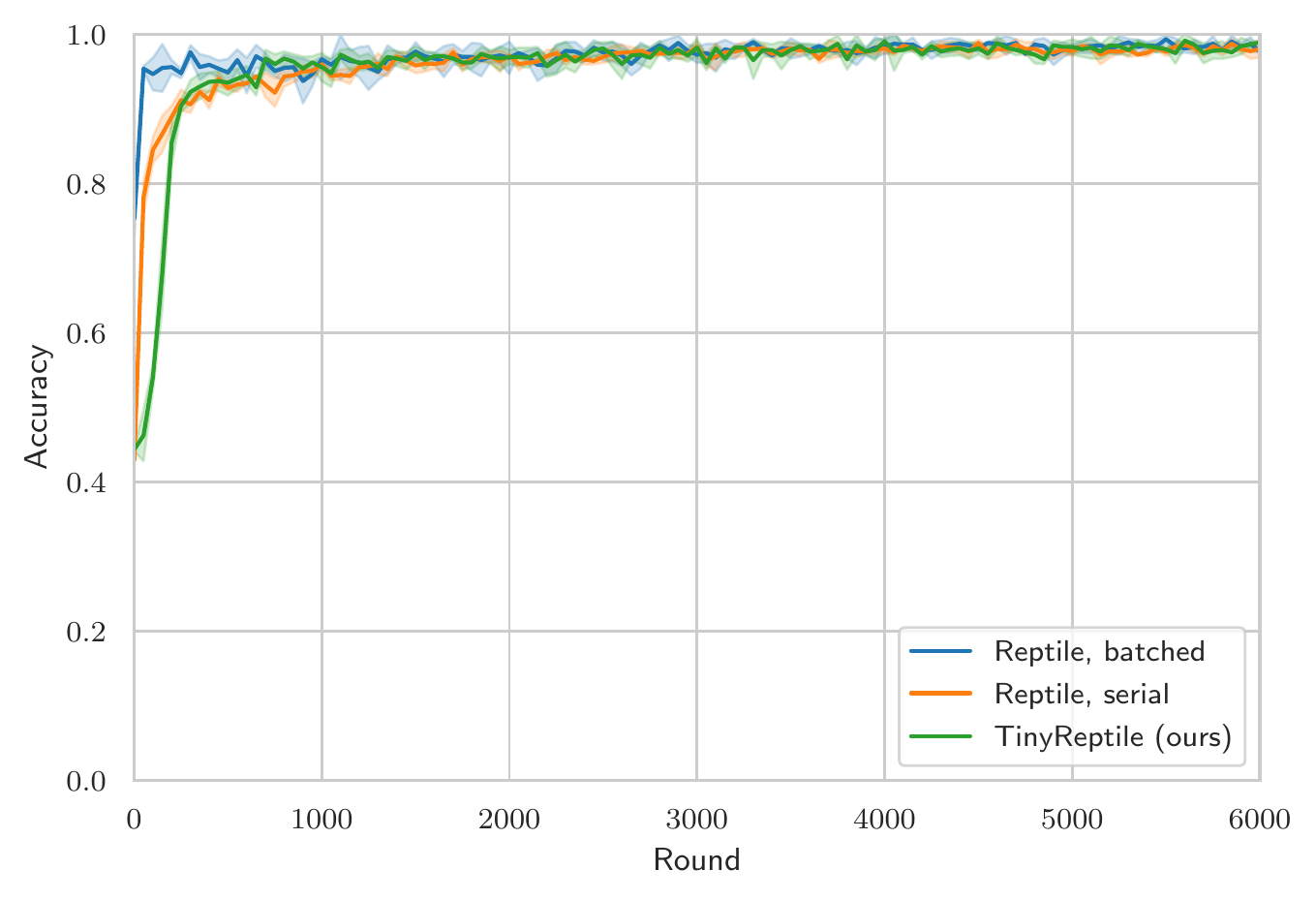}}

    \subfloat[Keywords spotting with four output classes.]{
    \includegraphics[width=1\columnwidth]{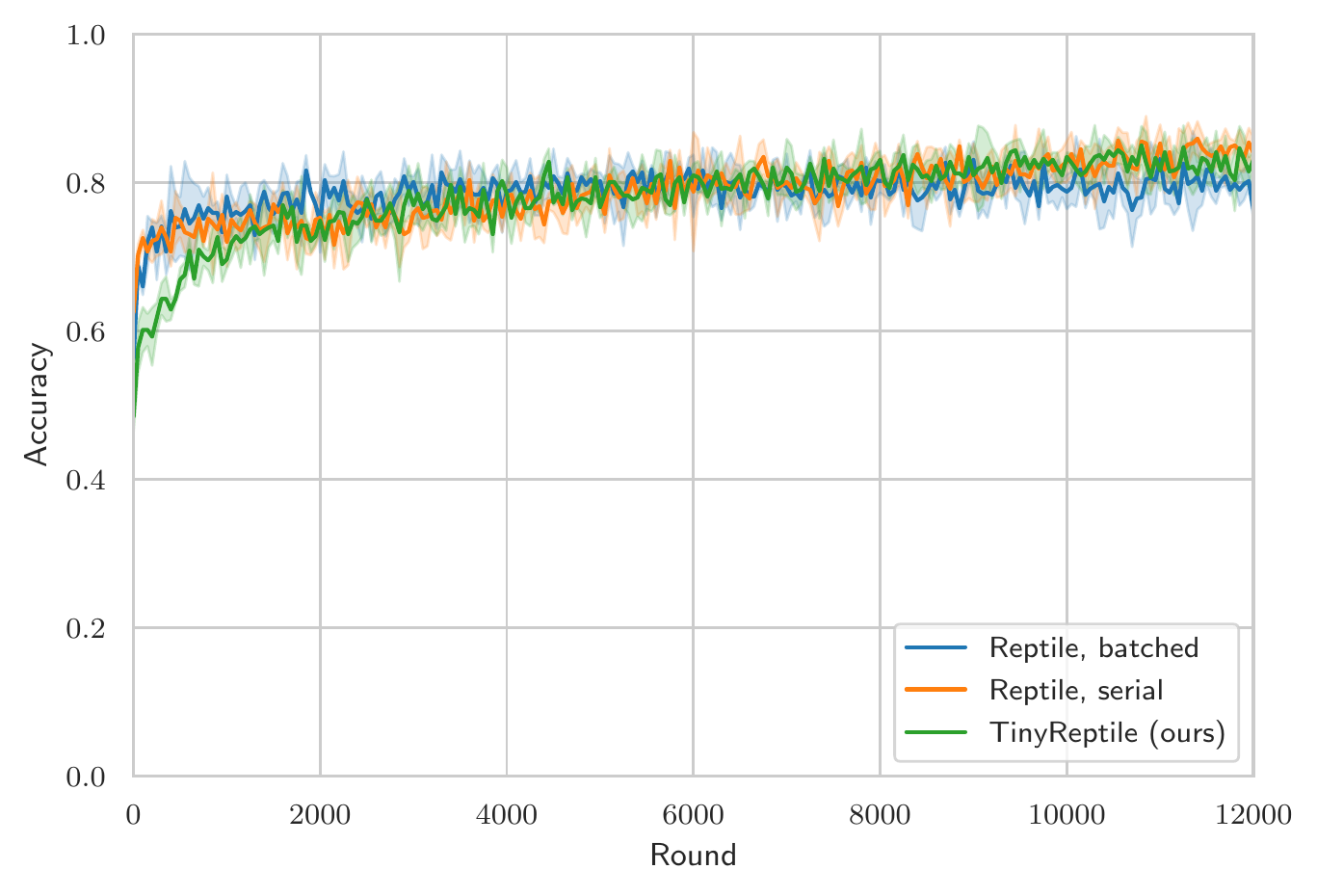}}
    
    \caption{Training convergence of Reptile  (batched \& serial) and TinyReptile on Omniglot and Keywords spotting datasets on Raspberry Pi 4. This shows that TinyReptile can achieve comparable performance to Reptile given sufficient training rounds.}
    \label{fig:4}
\end{figure}
 
In Fig.~\ref{fig:3}, we present the convergence performance of Reptile (serial) and TinyReptile on the Sine-wave example, where we also demonstrate the feasibility of running TinyReptile on the Arduino MCU. Arduino Nano 33 BLE Sense MCU is equipped with a 64-MHz Cortex-M4 CPU, 256-KB memory, and 1-MB flash, whose specification is within the typical range of TinyML devices. We choose the following hyperparameters in the experiments: $S_{training} = 32$, $\beta = 0.01$, and local epoch $E = 8$ (for training with Reptile). We observe a slower convergence, more fluctuations, and slightly worse results on the Arduino MCU than on the Raspberry Pi 4, which may be due to the limited numerical precision of the hardware. Nevertheless, given a sufficient number of training rounds, the model can still achieve reasonable performance on the MCU. In particular, Reptile (serial) performs worse than TinyReptile on the MCU. Our explanation is that because of the limited numerical precision of the MCU, the effect of the gradient calculated over the batch data may be canceled out and make the already less precise gradient even less precise.

Because the Arduino board is constrained for the Omniglot task and Keywords spotting task, we compare the training progress of Reptile and TinyReptile on these two datasets only on Raspberry Pi 4 in Fig.~\ref{fig:4}. Here, we choose the following hyperparameters in the experiments: $S_{training} = 16$, $\beta = 0.002$, local epoch $E = 8$ (for training with Reptile), and the number of sampled clients in each round $T = 32$ (for training with the batched version of Reptile). The results confirm that TinyReptile requires slightly more training rounds to obtain the same accuracy as Reptile, which is reasonable because TinyReptile updates on only one data point at a time, whereas Reptile trains on the entire support set in a batch. 

\begin{table}[tbp]
\centering
\caption{Comparison of time consumption of one training round on the Sine-wave example on Arduino Nano 33 BLE Sense (the results are measured under the support set size $S=32$).}
\resizebox{\columnwidth}{!}{%
\begin{tabular}{@{}lllll@{}}
\toprule
     & Sending & Local Training & Receiving & Total \\ \midrule
Reptile & 1.58 s    & 8.32 s          & 1.65 s     & 11.55 s \\
TinyReptile (ours) & 1.58 s   & 0.44 s          & 1.65 s     & 3.67 s \\ \bottomrule
\end{tabular}%
}
\label{tab:3}
\end{table}

\begin{table}[tbp]
\centering
\caption{Comparison of time consumption of one training round on Raspberry Pi 4 (the results are measured under the support set size $S=32$).}
\resizebox{\columnwidth}{!}{%
\begin{tabular}{@{}lll@{}}
\toprule
                           & \begin{tabular}[c]{@{}l@{}}Reptile\\ (batched \& serial)\end{tabular} & \begin{tabular}[c]{@{}l@{}}TinyReptile\\ (ours)\end{tabular} \\ \midrule
Sine-wave example          & 0.56 s                    & 0.24 s               \\
Keywords spotting (4 Classes) & 11.96 s                 & 3.45 s             \\
Omniglot (5 Classes)       & 22.53 s                 & 10.11 s             \\ \bottomrule
\end{tabular}%
}
\label{tab:4}
\end{table}

\subsubsection*{Time consumption}

We investigate the training time consumption of Reptile and TinyReptile to demonstrate the efficiency of TinyReptile. The results in Tables.~\ref{tab:3} and~\ref{tab:4} show that TinyReptile outperforms Reptile across all tasks in terms of training speed. In particular, on the constrained Arduino board, a 16X local training time reduction is achieved, indicating that TinyReptile is a perfect candidate for TinyML.
 
Lastly, we provide interested readers with the analysis of different hyperparameters in TinyReptile in Appendix~\ref{appendix:A} to support the easy deployment of TinyReptile.

\section{Conclusions and Future Work}
\label{section:conclusion}

According to Prof. Reddi and Pete Warden, the future of ML is bright and tiny~\footnote{\url{https://pll.harvard.edu/course/future-ml-tiny-and-bright?delta=0}}. Meta-learning is appealing for TinyML applications because we frequently require a customized ML solution for each TinyML device due to the heterogeneous deployment environment, and the number of MCUs keeps increasing. However, these constrained IoT devices are typically bare-metal devices without an operating system, making on-device learning an open challenge. Most TinyML solutions only support inference on them. This study presents TinyReptile, a lightweight meta-learning algorithm modified from the Reptile algorithm and tailored for TinyML scenarios. Building on the concepts of FL and online learning, TinyReptile can leverage distributed IoT devices with limited resources to collaboratively learn a solid NN initialization that can subsequently be quickly adapted to a new device/environment. Our contributions are as follows:

\begin{itemize}
\item We propose TinyReptile, which, to the best of our knowledge, is the first attempt to implement meta-learning on resource-constrained devices.
\item We introduce the audio classification dataset “Keywords spotting,” derived from the “Speech command” dataset, enlightening a new classification problem for evaluating meta-learning in the context of TinyML.
\item We evaluate TinyReptile for various meta-learning tasks, compare it with other approaches, and measure its real-world performance on Raspberry Pi 4 and Arduino Nano BLE 33 Sense MCU. The results confirm that TinyReptile can achieve comparable performance to the baseline algorithms while saving at least 50\% in resource consumption and training time.
\item We study the impact of several hyperparameters in TinyReptile, as illustrated in Appendix~\ref{appendix:A}, to guide its real-world deployment. 
\end{itemize}

Future work already underway includes further improvement of the algorithm, such as applying learning rate annealing techniques, providing best practices in real-world deployment by fine-tuning hyperparameters, benchmarking on other meta-learning tasks, comparing the algorithm with other state-of-the-art approaches, extending the evaluations to other tiny devices, applying the algorithm to large-scale industrial use cases, and investigating the impact of the reduced numerical precision of MCUs.

\section*{Acknowledgment}

This work is partially supported by the NEPHELE project (ID: 101070487) that has received funding from the Horizon Europe programme under the topic "Future European platforms for the Edge: Meta Operating Systems".

\bibliographystyle{IEEEtran}
\bibliography{conference_101719}

\appendix

\subsection{Recipe for real-world deployment}
\label{appendix:A}

\begin{figure}[bbp]
\centerline{\includegraphics[width=1\columnwidth]{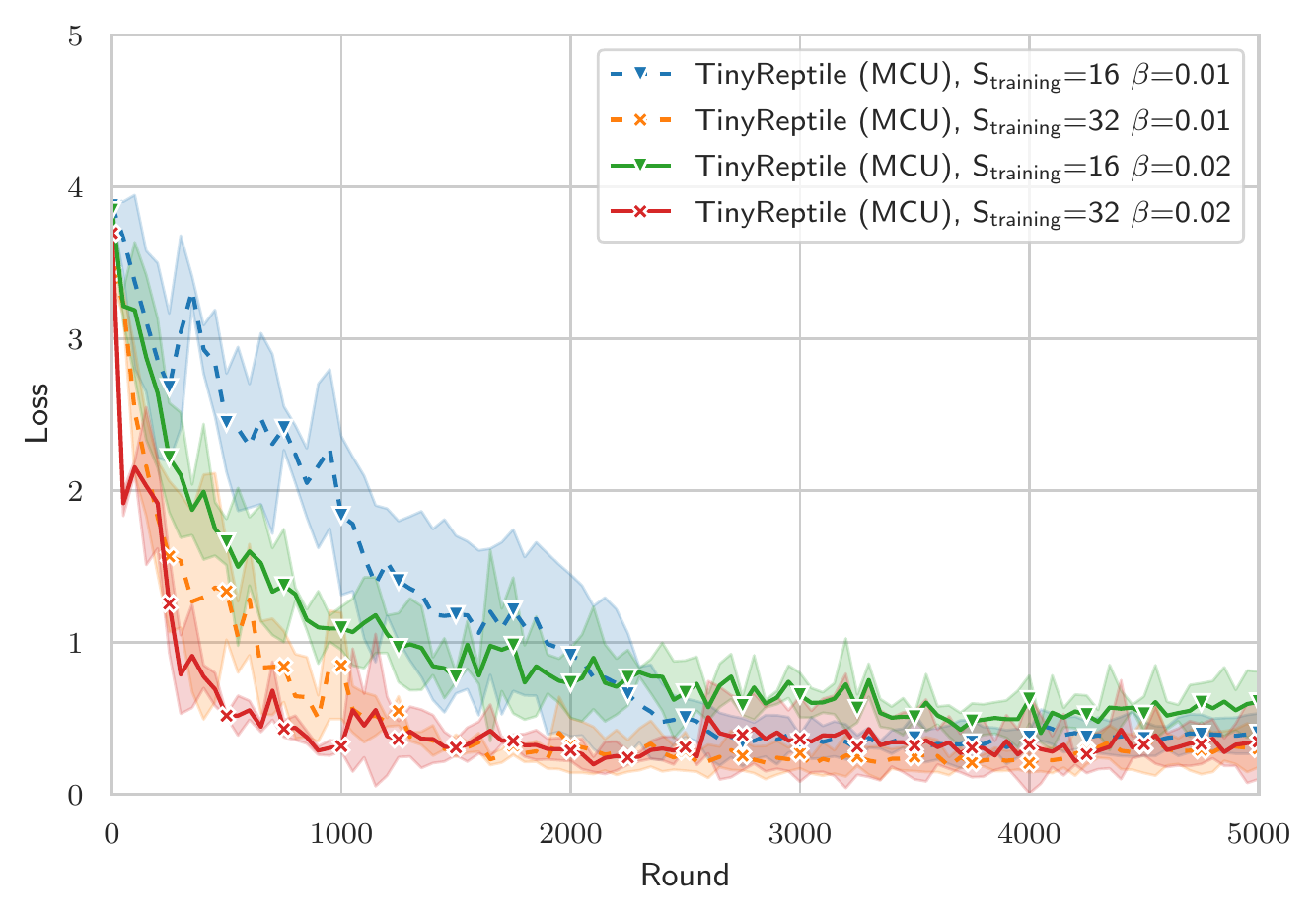}}
\caption{Training convergence of TinyReptile on the Sine-wave regression example on the Arduino Nano 33 BLE Sense MCU.}
\label{fig:5}
\end{figure}

Here, we study how the ingredients, i.e., hyperparameters, impact the performance of TinyReptile and provide guidance for the real-world deployment of TinyReptile.

First, we investigate the effect of $\beta$ and $S_{training}$ on the Sine-wave example on the Arduino board. The results in Fig.~\ref{fig:5} show that larger support set size of training clients $S_{training}$ can result in improved performance. Besides, a higher client learning rate $\beta$ can contribute to faster convergence in the early training phase but not necessarily to better final performance. This suggests that a learning rate annealing technique can be helpful in practice. 

\begin{figure}[tbp]
    \centering
    \subfloat[Sine-wave example.]{
    \includegraphics[width=1\columnwidth]{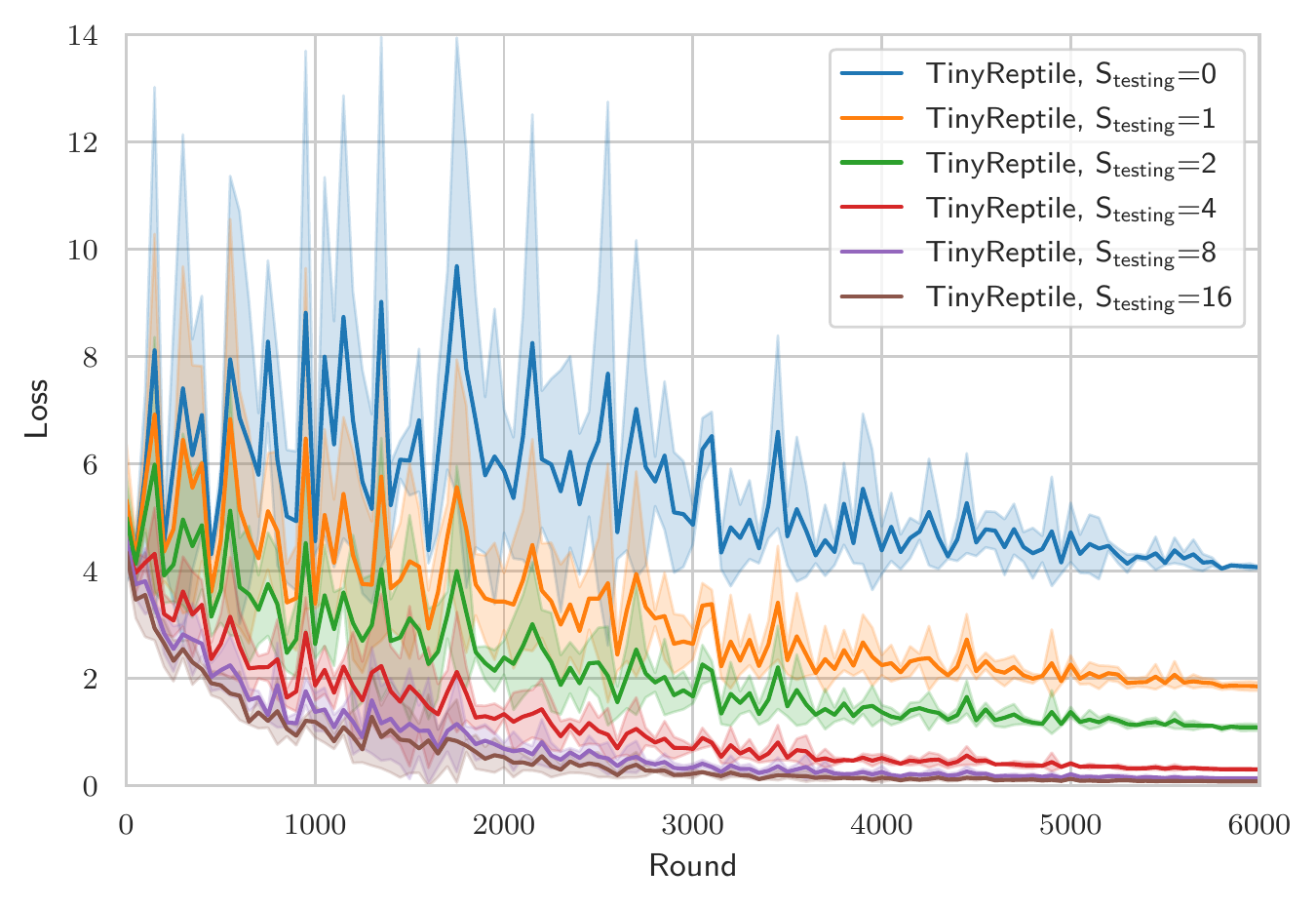}}
    
    \subfloat[Omniglot with five output classes.]{
    \includegraphics[width=1\columnwidth]{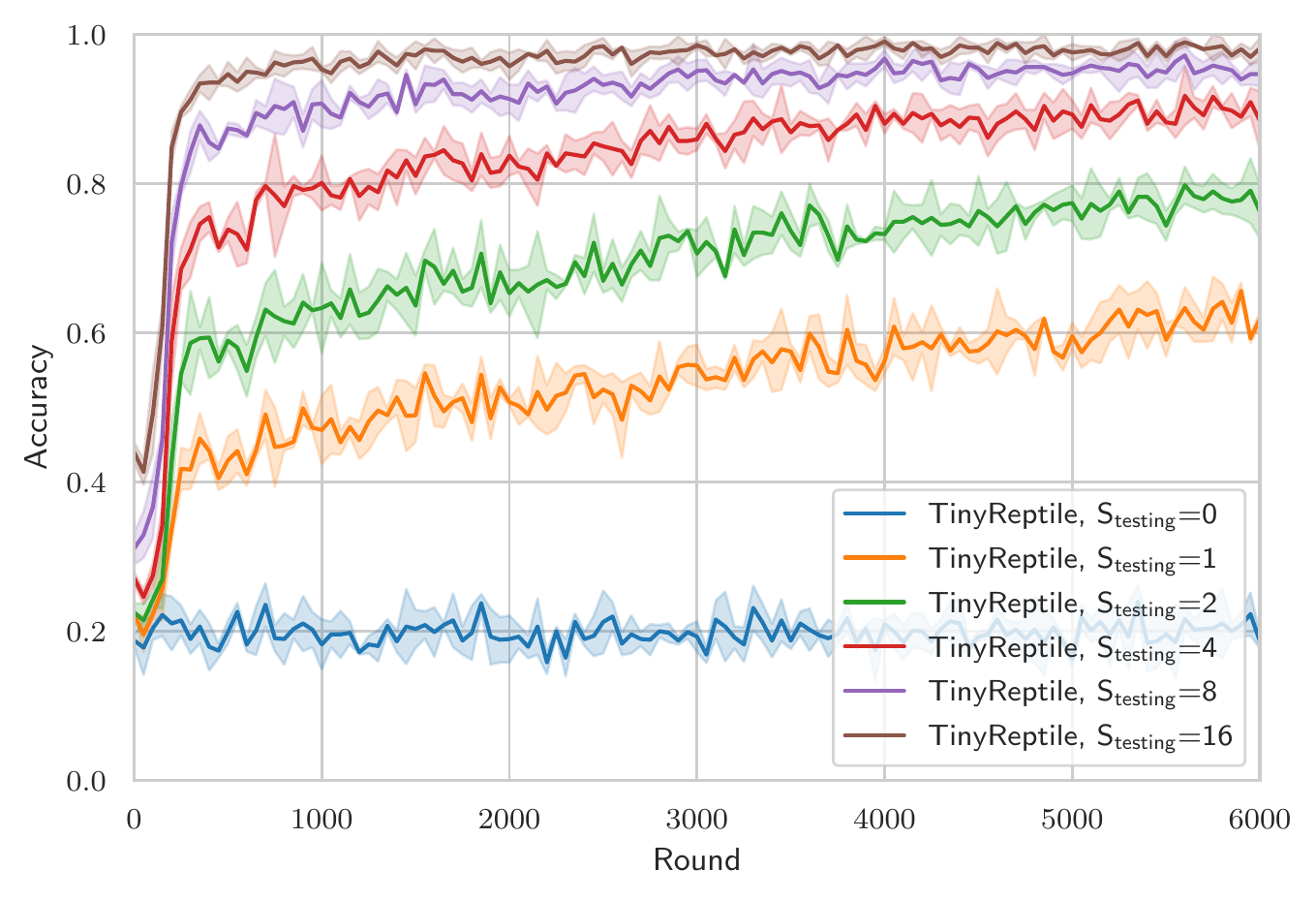}}

    \subfloat[Keywords spotting with four output classes.]{
    \includegraphics[width=1\columnwidth]{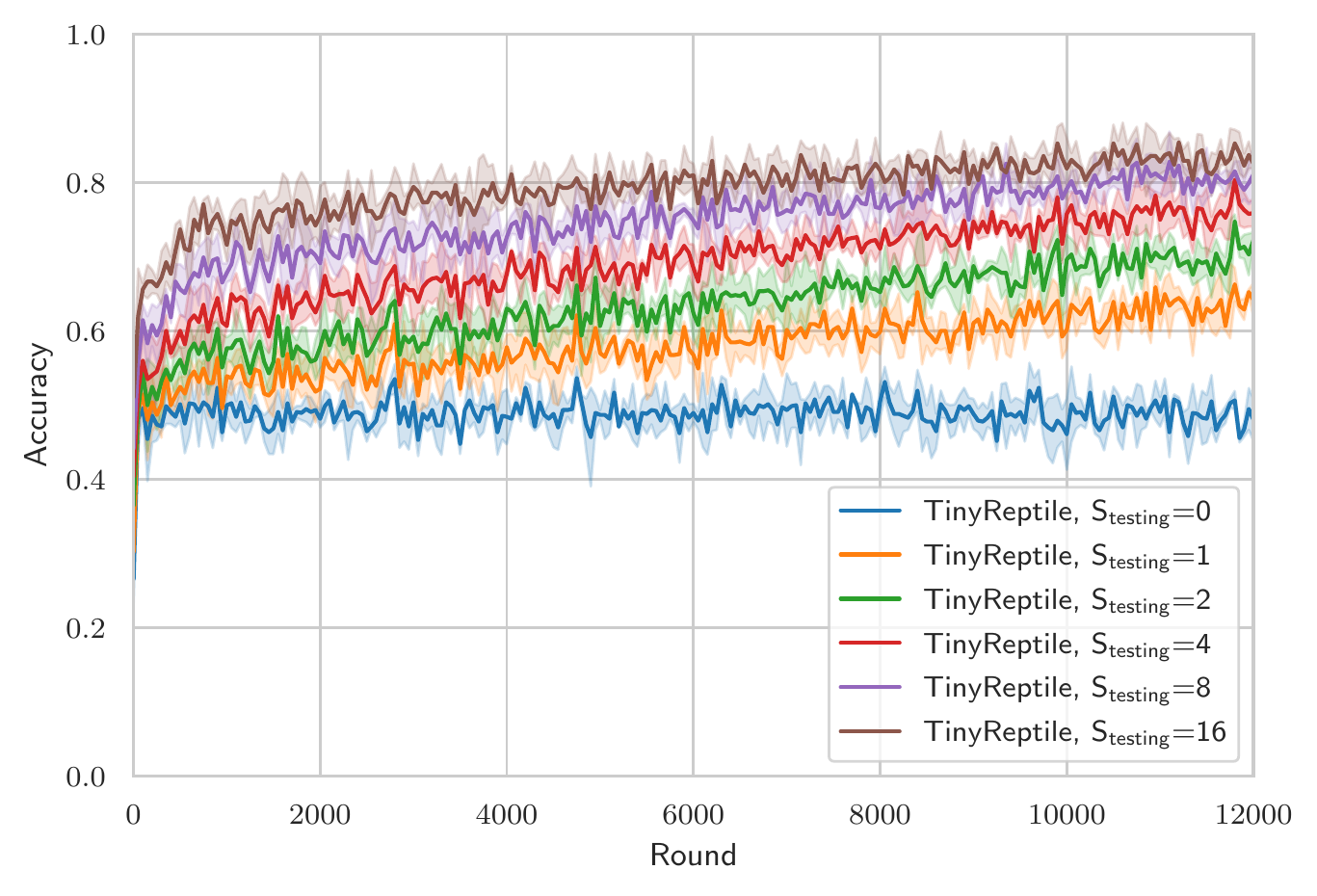}}
    
    \caption{Testing accuracy of TinyReptile as a function of support set size S\textsubscript{testing} of testing clients on the Sine-wave example, Omniglot, and Keywords spotting datasets.}
    \label{fig:6}
\end{figure}

Next, we discuss the effect of the support set size of testing clients $S_{testing}$ because TinyML clients frequently have very limited or even no labeled data for local fine-tuning. Specifically, we want to investigate how much data are necessary for local fine-tuning to achieve good performance. As depicted in Fig.~\ref{fig:6}, the global initialization trained with TinyReptile does not generalize well if no support data are available. However, a significant improvement can be achieved when we provide the model with even one data pair for adaptation. Moreover, the larger the $S_{testing}$, the better the performance.

\vspace{12pt}

\end{document}